\pdfoutput=1

\documentclass[11pt]{article}

\usepackage[final]{acl}
\usepackage{times}
\usepackage{latexsym}

\usepackage[T1]{fontenc}

\usepackage[utf8]{inputenc}

\usepackage{microtype}

\usepackage{inconsolata}

\usepackage{svg}  
\usepackage{multirow}
\usepackage{float}
\usepackage{CJK}
\usepackage{graphicx}
\usepackage{algpseudocode}
\usepackage{amsmath}
\usepackage{soul}
\usepackage{xcolor}
\usepackage[switch]{lineno}
\usepackage{multirow}
\usepackage{float}
\usepackage{CJK}
\usepackage{xcolor}
\usepackage{times}
\usepackage{soul}
\usepackage{url}
\usepackage{caption}
\usepackage{booktabs}
\usepackage[switch]{lineno}
\usepackage{svg}  
\usepackage{amssymb}
\usepackage{multirow}
\usepackage{float}
\usepackage{CJK}
\usepackage{subcaption}
\usepackage{algorithm,algpseudocode}
\usepackage{amsmath}
\usepackage{soul}
\usepackage{xcolor}
\usepackage{url}
\usepackage{booktabs}
\usepackage[switch]{lineno}
\usepackage{multirow}
\usepackage{float}
\usepackage{listings}
\title{H-MEM: Hierarchical Memory for High-Efficiency \\Long-Term Reasoning in LLM Agents}

\author{Haoran Sun,  Shaoning Zeng} 

\begin{document}
\maketitle
\begin{abstract}
Long-term memory is one of the key factors influencing the reasoning capabilities of Large Language Model Agents (LLM Agents). Incorporating a memory mechanism that effectively integrates past interactions can significantly enhance decision-making and contextual coherence of LLM Agents. While recent works have made progress in memory storage and retrieval, such as encoding memory into dense vectors for similarity-based search or organizing knowledge in the form of graph, these approaches often fall short in structured memory organization and efficient retrieval. To address these limitations, we propose a Hierarchical Memory (H-MEM) architecture for LLM Agents that organizes and updates memory in a multi-level fashion based on the degree of semantic abstraction. Each memory vector is embedded with a positional index encoding pointing to its semantically related sub-memories in the next layer. During the reasoning phase, an index-based routing mechanism enables efficient, layer-by-layer retrieval without performing exhaustive similarity computations. We evaluate our method on five task settings from the LoCoMo dataset. Experimental results show that our approach consistently outperforms five baseline methods, demonstrating its effectiveness in long-term dialogue scenarios.
\end{abstract}

\section{Introduction}

Large Language Models (LLM) Agents possess the capability to autonomously make decisions and perform a wide range of tasks, with particularly strong performance in question-answering scenarios \cite{agentsurvey,LLMsurvey1}. In long-term conversational settings, the decision-making and reasoning processes of LLM Agents often require integrating historical interactions, thereby relying heavily on their internal memory mechanisms \cite{agentsurvey2,LLMsurvey2}. The memory mechanism is designed to emulate the dynamic nature of human-like cognitive memory by retaining prior dialogue information \cite{LLMsurvey3}. This allows the model to effectively retrieve relevant memory content during response generation, enabling context-aware and personalized replies \cite{memorysurvey}. The effective design of memory mechanisms plays a critical role in enhancing the performance of LLM Agents in complex and extended dialogue tasks \cite{LLMsurvey5}.

\begin{figure}[t]
    \centering
    \includegraphics[width=1\linewidth]{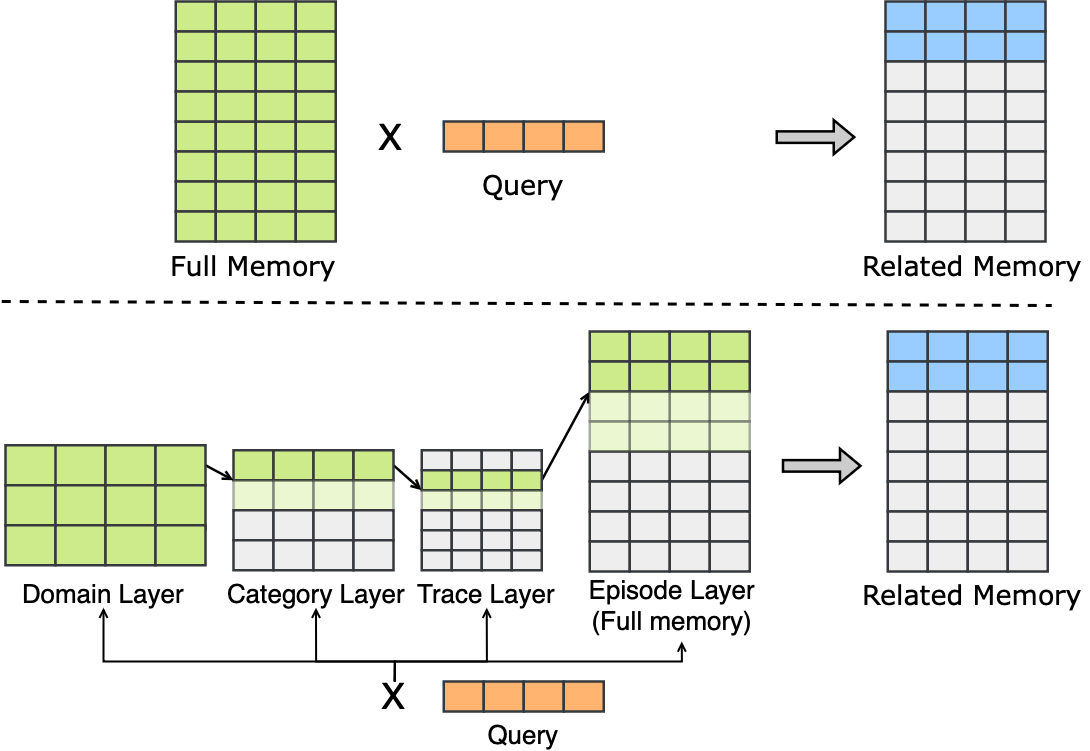}
    \caption{\textbf{Memory Architecture Comparison. }The above is the traditional memory mechanism, which make query to calculate the similarity with all stored specific memories and selects the top-k related memories \cite{memorybank}. The following is H-MEM, which uses hierarchical memory and position index to search layer by layer and can effectively remove the influence of irrelevant memories on calculation.}
    \label{fig:intro}
\end{figure}

The most basic memory mechanism involves simple multi-turn dialogue, where all previous interactions with LLM are concatenated with the current prompt and user query \cite{LLMsurvey6, multidialogue}. However, this approach is limited by the context window length of LLMs, making it ineffective for long-term interactions. This limitation has promoted researchers to explore more efficient memory mechanisms. MemGPT adopts a layered memory architecture inspired by operating systems, combining a limited context window with external storage \cite{memgpt}. However, its external memory design relies on retrieval-augmented models, which may face trade-offs between retrieval accuracy and response efficiency when dealing with large-scale information. MemInsight strengthens the memory representation of LLM agents by autonomously extracting key information and generating structured attributes \cite{meminsight}. A-MEM, inspired by the Zettelkasten method, implements a self-organizing and evolving memory system by dynamically constructing a network of knowledge \cite{Amem}. Nonetheless, the complexity of such memory structures introduces consistency and accuracy challenges in maintaining and updating inter-node relationships. MemoryBank enhances long-term interaction capabilities by encoding memories into vector representations and retrieving them through similarity-based search \cite{memorybank}. This vector-based encoding approach improves the efficiency of both memory storage and retrieval \cite{memorysurvey}. Nevertheless, as the number of memory entries increases, the computational complexity of vector retrieval also rises sharply, posing challenges to system performance.

In order to optimize the ability of existing memory mechanisms in structured storage, systematic organization, and efficient retrieval, we propose a novel Hierarchical Memory (H-MEM) architecture. H-MEM adopts a hierarchical memory structure, dividing memory into four levels according to the degree of semantic abstraction, corresponding to structures similar to section, subsection, subsubsection, and content. In this structure, the memory vectors of each level are embedded in the position index encoding of their subordinate sub-memories, which enables efficient screening of relevant memory units through upper-level semantic information during the retrieval process, and then continues to retrieve the corresponding related sub-memories of the next layer based on the relevant memories. As shown in Figure \ref{fig:intro}, H-MEM not only enhances the organization and structure of memory retrieval, but also significantly reduces the participation range of irrelevant information, thereby effectively reducing computational costs and achieving efficient and targeted memory access. Our main contributions are summarized as follows.

\begin{itemize}
\item H-MEM, a hierarchical memory architecture for LLM agents, is proposed to integrate multi-level memory storage with positional index encoding of sub-memory at each layer, enabling structured and systematic memory organization as well as efficient and orderly memory retrieval. In addition, we design a more scientific memory update mechanism to adapt to the complex psychological changes of humans.

\item We compare H-MEM with five baseline methods across multiple large language models on five question-answering tasks from the LoCoMo dataset. Experimental results demonstrate the effectiveness of H-MEM in optimizing memory storage structures and the efficiency of memory retrieval.

\end{itemize}

\begin{figure*}[t]
    \centering
    \includegraphics[width=1\linewidth]{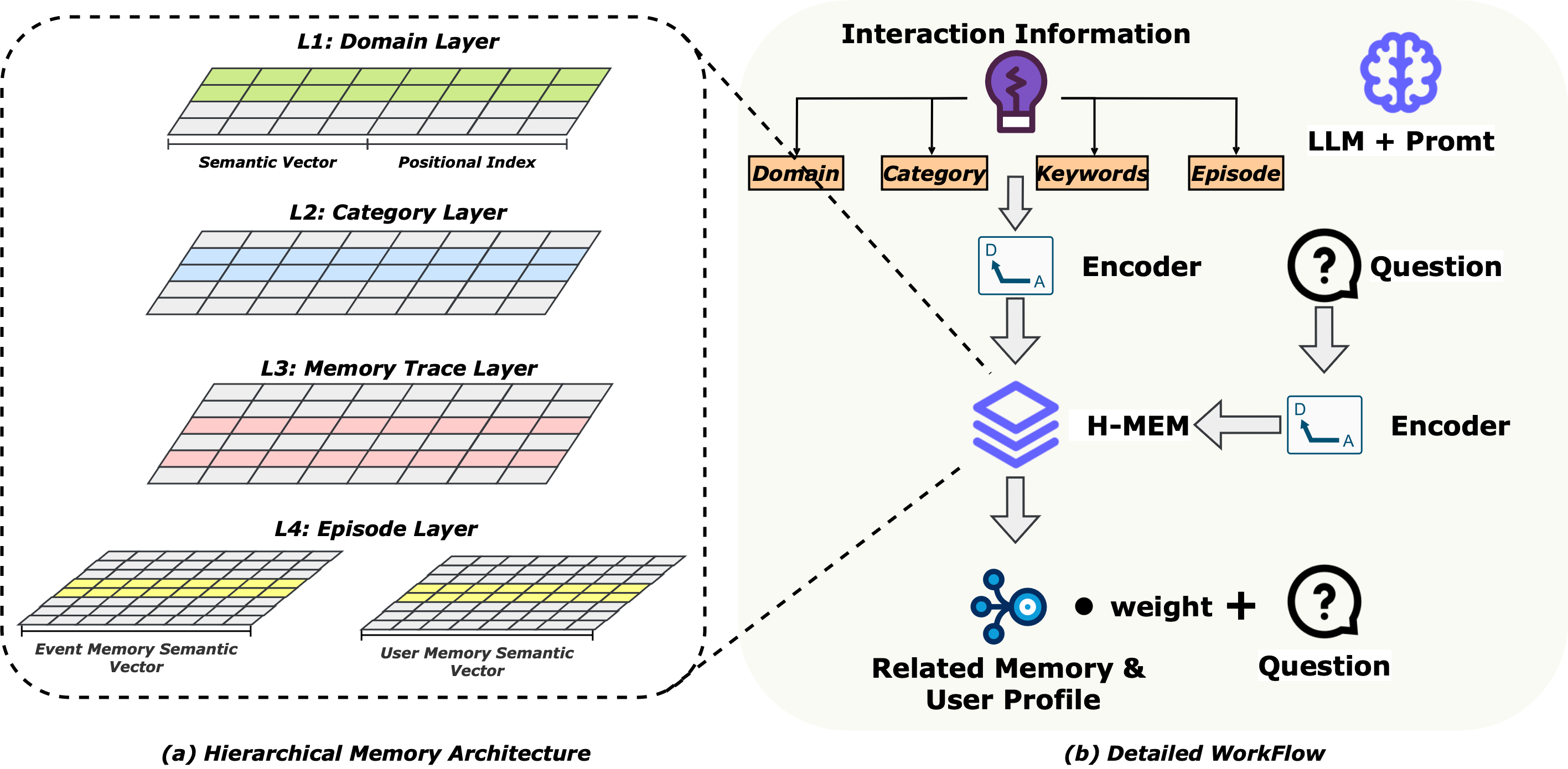}
    \caption{\textbf{H-MEM architecture. }\textbf{(a)} shows the hierarchical memory structure of H-MEM, divided into four memory layers: Domain Layer, Category Layer, Memory Trace Layer, and Episode Layer. \textbf{(b) }shows the specific memory extraction workflow of H-MEM. After encoding the questions into semantic vectors, perform similarity calculation with H-MEM memory, select the most relevant top-k related memories and user profile, attach corresponding memory weights to provide LLM with Confidence Level reference, and input them together with the questions to LLM to achieve long-term dialogue reasoning.}
    \label{fig:architecture}
\end{figure*}

\section{Related Work}
\subsection{Memory for LLM Agents}
To enhance the long-term reasoning capabilities of LLM Agents, numerous researchers have explored memory mechanisms of it \cite{memorysurvey1,agentsurvey,needmemory}. ReadAgent is an LLM-based agent system that significantly extends the effective context length of LLMs and improves reading comprehension by segmenting long texts into pages, compressing them into key-point memories, and performing interactive retrieval when needed \cite{readingagent}. MemGPT leverages a virtual context management technique to extend the context window of LLMs \cite{memgpt}. It adopts a layered memory architecture similar to that of an operating system, integrating limited internal context with external storage. Through function calls, it dynamically manages data, allowing effective processing of long documents and multi-turn dialogues \cite{memorysurvey1}. The Self-Controlled Memory (SCM) framework strengthens LLMs' ability to handle long texts and long-term memory via three core components: the LLM agent, memory stream, and memory controller \cite{SCM}. SCM dynamically determines when and how to utilize memory, improving information retrieval and response quality. A-MEM, inspired by the Zettelkasten method, is a memory system for LLM agents that enables self-organizing and evolving memory through the dynamic construction of a knowledge network. It creates structured memory notes for new entries and establishes connections with historical memories through contextual analysis. As new memories are added, the system updates contextual representations of existing ones, progressively optimizing the knowledge structure and enhancing LLM agent performance in long-term interaction tasks \cite{Amem}. MemInsight boosts the representational power of memory of LLM Agents by autonomously extracting key information and generating attributes \cite{meminsight}. This approach significantly improves semantic data retrieval efficiency and accuracy across tasks such as question answering, conversational recommendation, and event summarization. 

MemoryBank introduces a long-term memory mechanism for large language models, enhancing their long-term interaction capabilities through memory storage, retrieval, and update strategies inspired by the Ebbinghaus forgetting curve \cite{memorybank}. This mechanism enables effective recall of past information, better user understanding, and more personalized and natural interactions \cite{memorybank}. While this method solves the long-term memory problem of LLM Agents well, there are still shortcomings in structured memory storage and efficient systematic memory retrieval, and further optimization and improvement are needed.

\section{Method}

\subsection{Memory Storage}



The storage layer of H-MEM is organized into a four-level hierarchical structure, designed according to increasing levels of semantic abstraction and generalization. As shown in Figure \ref{fig:architecture}, from top to bottom, these layers are: Domain Layer, Category Layer, Memory Trace Layer, and Episode Layer. This hierarchy resembles the structure of a document, analogous to section, subsection, subsubsection, and content, respectively. The first three layers serve as a progressively refined index, providing a systematic and interpretable organization of memory, while the bottom layer contains the actual episodic content and user profile information. The approximate simplified meaning of the prompt is: \textit{"You are a information analyze agent for a long-term LLM system. Given a dialogue, you must extract and structure the information into a hierarchical memory format. Follow this hierarchy strictly: 1. Identify the high-level domain of interest. 2. Extract specific categories or subdomains related to the topic. 3. Summarize the keywords of the dialogue. 4. Extract specific events and user profile. Output the result as structured JSON. "}.

After each interaction between the user and the large language model (LLM), a specialized memory extraction model is invoked to analyze the interaction and extract multi-level structured information. Guided by carefully designed prompts, this model parses the interaction into four semantic layers. For example, when a user requests a recommendation for an action movie or a skiing competition, the LLM might suggest a Kung Fu movie starring Jackie Chan and a competition featuring Mikaela Shiffrin. 
The first three layers store abstract summaries similar to directories. Episode Layer stores the complete contextual memory of the interaction, including a timestamp and an inferred user profile. This profile reflects the user’s preferences, interests, emotional states, and behavioral patterns. All memory entries are encoded into dense vector representations using a neural encoder to support efficient semantic retrieval. The structured design of H-MEM enhances the interpretability of stored information while ensuring scalability for memory retrieval in long-term interactions, thus enabling more fine-grained reasoning. In the Episode Layer, both vector representations and textual memory are preserved: the vectors are used for similarity computation, while the text is used to select the final memory content, which is then integrated with prompts to provide accurate memory grounding for the LLM.

We also designed a self-adaptation hierarchy adjustment interface, allowing users to dynamically adjust the hierarchy structure based on the complexity of the current conversation and the semantic granularity of the memorized content. For example, in simple conversations, the number of levels can be reduced to improve efficiency; while in conversations involving multiple topics and complex relationships, levels can be added to better organize memory.

\subsection{Memory Retrieval}

After storing the memory vector of an interaction into different layers of memory, H-MEM will embed self position index encoding after each memory in each layer, and embed the position index encoding corresponding to all subordinate layer sub-memories after the first three layers of memory. Formally, for a memory entry at layer $L$, denoted as $\mathbf{v}_i^{(L)}$, its representation is defined as:

\[
\mathbf{v}_i^{(L)} = \left[ \underbrace{\mathbf{e}_i^{(L)} \in \mathbb{R}^D}_{\text{Semantic Vector}},\underbrace{p_{(i-1)x}}_{\text{Self Index}}, \underbrace{p_{i1}, \ldots, p_{iK}}_{\text{Sub-Memories Indices}} \right]
\]

Here, $\mathbf{e}_i^{(L)}$ is a dense semantic vector capturing the high-level meaning of the memory entry at layer $L$, while $p_{(i-1)x}$ is the position index of the memory itself and $p_{i1}, \ldots, p_{iK}$ are discrete position indices pointing to its semantically related sub-concepts in the next layer ($L+1$), allowing efficient index-based routing without the need for exhaustive similarity computation. During inference, a top-down memory traversal is performed. 

The query is first embedded and calculated for similarity with the semantic vectors $\mathbf{e}_i^{(L)}$ at the highest abstraction layer. In H-MEM, we choose to use the FAISS library for calculation to achieve more efficient similarity calculation. When calculating similarity, the positional index encoding is ignored and only calculated between the semantic vectors of the memory content. Once the top-k relevant memory entry is selected, the associated index pointers $\{p_{i1}, \ldots, p_{iK}\}$ guide the retrieval to the corresponding entries in layer $L+1$. Among them, $i$ represents which layer of memory it belongs to, and the number after $i$ represents the number of rows in the $i-th$ layer of memory. Specifically, the memories are hierarchically organized into $L$ levels: $\mathcal{M}^{(1)}, \mathcal{M}^{(2)}, \dots, \mathcal{M}^{(L)}$, where level $l$ contains memory units $\mathcal{M}^{(l)}$, and each unit in level $l$ contains child entries in level $l+1$. Let $q$ be the query vector. The top-k retrieval at each level is defined recursively as:

\[
\mathcal{M}^{(l)}_k = \bigcup_{x \in \mathcal{M}^{(l-1)}_k} \operatorname{TopK}_{y \in \text{Child}(x)} \left( \operatorname{sim}(q, y) \right)
\]

This process can be recursively applied until the most fine-grained memory is reached. Finally, each selected memory will be accompanied by its own memory weight, providing a Confidence Level reference for LLM. After calculating and experimenting with different L values, we finally chose the optimal 4-layer, which can simultaneously balance the accuracy and efficiency of retrieval.

To facilitate a fair comparison of computational complexity between traditional retrieval methods and the proposed H-MEM framework, we define a set of fixed conditions. Suppose there are $a$ domains in total. Each domain contains 100 categories, each category includes 100 memory traces, and each trace comprises 100 episodes. Therefore, the total number of fine-grained memory entries is $a \cdot 100 \cdot 100 \cdot 100 = a \cdot 10^6$. Each memory entry is represented by a vector of dimension $D$. Existing approaches store all memory entries in a flat structure and perform retrieval directly over the entire memory set, resulting in a computational complexity of 
$\mathcal{O}(a \cdot 10^6 \cdot D)$. In contrast, H-MEM adopts a hierarchical retrieval strategy. It first selects the top-$k$ most relevant domains from the $a$ candidates. For each selected domain, it then retrieves the top-$k$ most relevant categories (from $k \cdot 100$ candidates), followed by retrieving $k$ memory traces from $k \cdot 100$ trace candidates. Finally, it selects the top 10 most relevant episodes from $k \cdot 100$ episodes. The overall computational complexity of H-MEM can be approximated as $\mathcal{O}((a + k \cdot 300) \cdot D)$. This hierarchical retrieval not only significantly reduces the computation cost compared to the traditional memory retrieval method but also ensures structured, layer-wise filtering of memory, thereby enhancing both relevance and accuracy.

\begin{figure}[t]
    \centering
    \includegraphics[width=1\linewidth]{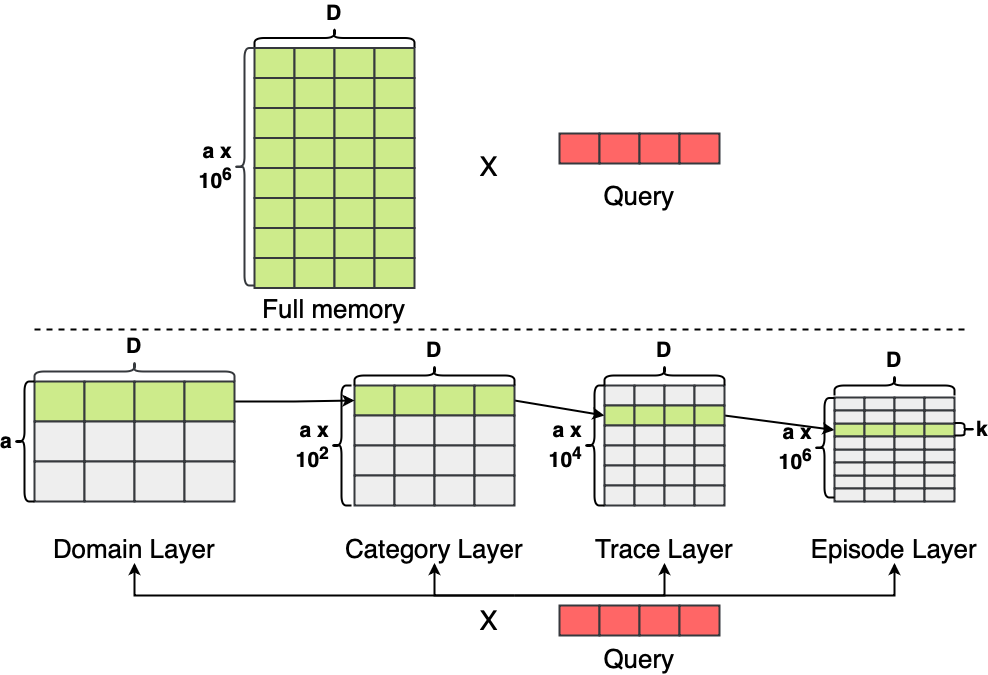}
    \caption{\textbf{Memory Retrieval Calculation Comparison. }The above is the traditional memory retrieval method \cite{memorybank}, which make query to calculate the similarity with all stored specific memories and selects the top-k related memories. The following is H-MEM, which uses position index to search layer by layer.}
    \label{fig:compare}
\end{figure}

\subsection{Memory Update}

Existing research has effectively simulated the forgetting process of human memory with the help of the Ebbinghaus Forgetting Curve, taking into account the characteristics of the enhanced strength of memory after it is called upon \cite{memorybank}. However, such methods ignore the dynamic changes in human psychological states. In fact, human interests and behavioral preferences are complex and changeable. For example, a user may have been enthusiastic about skiing before, but after a period of time, due to the influence of some intermediate events, they developed a strong aversion to skiing. To more realistically model the variability of human memory, we introduce a dynamic memory regulation mechanism based on LLM based on user feedback on the basis of traditional forgetting curves. Specifically, when LLM calls a certain memory to generate an answer, H-MEM will adjust the weight of the corresponding memory based on the feedback performance (approval, no obvious feedback, or rebuttal) of user: if the user shows approval, the weight of the memory will be enhanced, which is regarded as the reinforcement of effective memory; if there is no feedback, it will follow the original forgetting curve Natural Shrink; if the user refutes, the weight of the memory will be reduced, indicating that the memory may have expired. The above enhancement or weakening process is updated by multiplying with the feedback weight generated by LLM, thereby achieving dynamic adjustment of memory strength.
 
\section{Experiment}

\subsection{Setup}
\noindent \textbf{Datasets. }To evaluate the effectiveness of H-MEM in long-term dialogue settings, we adopted the LoCoMo dataset following established practices. LoCoMo dataset is specifically designed to assess the memory capabilities of large language model (LLM) agents in extended multi-session interactions. It comprises 50 dialogues, each averaging 300 turns and spanning up to 35 sessions, with an average of 9,000 tokens per dialogue. The dataset includes 7,512 question-answer pairs categorized into five types: (1) single-hop questions answerable from a single session (SH.)- 2705 pairs; (2) multi-hop questions requiring information synthesis across sessions (MH.)- 1104 pairs; (3) temporal reasoning questions testing understanding of time-related information (T.)- 1547 pairs; (4) open-domain knowledge questions requiring integration of conversation context with external knowledge (OD.)- 285 pairs; and (5) adversarial questions assessing models’ ability to identify unanswerable queries (A.)- 1871 pairs. LoCoMo ensures long-range consistency across entire dialogues, making it well-suited for evaluating a LLM’s ability to handle long-distance dependencies and maintain coherence over extended conversations.\\
\\

\noindent \textbf{Implementation Details. }We conduct experiments using Qwen-1.5B/3B \cite{qwen2}, LLaMA 3.2-1B/3B \cite{llama}, and DeepSeek-R1 1.5B/7B \cite{deepseek-r1} as the base question-answering models. BERT is employed as the encoder for memory storage. For similarity computation, we utilize FAISS \cite{faiss} and retrieve the top-k memories with k = 10. Regarding the baselines, we follow the original papers in terms of experimental data and configuration settings \cite{memgpt,memorybank,Amem}. DeepSeek-R1-8B \cite{deepseek-r1} is primarily used for memory analysis and information extraction. To eliminate the influence of model type, we also include Qwen3-8B \cite{qwen2.5} and LLaMA 3.2-3B \cite{llama} in ablation studies. All models used in our experiments are deployed locally via Ollama on a system equipped with two NVIDIA RTX 4090 GPUs.\\
\\

\noindent \textbf{Baselines. }In the experiments, we compare our work on the five types of tasks with other long-term memory methods of LLM. These methods include LoCoMo (LCM.) \cite{LoCoMoDataset}, ReadAgent (RA.) \cite{readingagent}, MemoryBank (MB.) \cite{memorybank}, MemGPT (MG.) \cite{memgpt}, and A-MEM (AM.) \cite{Amem}. Among them, the LoCoMo method is the original evaluation method of the LoCoMo dataset itself, similar to the general multi-round dialogue, without using special memory mechanisms, directly using the basic model for question answering tasks. Before each interaction, the complete previous historical interaction information is combined with the question and prompt.\\
\\

\noindent\textbf{Evaluation Metrics. }We utilize two primary metrics to comprehensively assess the performance of the LLM equipped with different memory. The F1 score is employed to evaluate the accuracy of the answers by calculating the harmonic mean of precision and recall, thereby providing a balanced measure of the model's ability to generate both relevant and complete responses. The BLEU-1 score is used to assess the quality of the generated responses by measuring word overlap with ground truth responses between the output of LLM, which helps in evaluating the lexical precision of the generated text. \\
\\

\noindent \textbf{Computational Efficiency Experiment Design. }In addition to the effectiveness evaluation experiments described earlier, we designed a Computational Efficiency Analysis experiment to assess the memory retrieval efficiency of H-MEM in a continuous reasoning scenario with huge amount of irrelevant memories. In this setting, five task types—Single Hop, Multi Hop, Temporal, Open Domain, and Adversarial—are executed sequentially as a single run. During the process, the stored memory is not cleared between task switches but is continuously accumulated. This design simulates real-world usage and aims to evaluate whether H-MEM can maintain efficient memory retrieval as the volume of stored memory increases. Among all the baseline methods, only MemoryBank and H-MEM adopt a memory mechanism that encodes memory into vector representations and retrieves it via similarity-based matching. Therefore, we conduct the efficiency comparison solely between H-MEM and MemoryBank. In this experiment, we use BERT as the encoder, employ Flat of FAISS for similarity-based retrieval, and set the top-k parameter to k=10. The base model for the tasks is Qwen-1.5B, while the memory reasoning module is implemented using DeepSeek-R1-8B.

\begin{table*}[h]
    \centering
\scriptsize
  
  \begin{tabular}{cc|c|cccccccccc|cc}
    \toprule  
   \multicolumn{2}{c|}{ \multirow{2}*{\textbf{Models}}}  &\multirow{2}*{\textbf{Baselines}}
    
    &\multicolumn{2}{c}{ \textbf{Single Hop}}&\multicolumn{2}{c}{ \textbf{Multi Hop}}&\multicolumn{2}{c}{ \textbf{Temporal}}&\multicolumn{2}{c}{ \textbf{Open Domain}}&\multicolumn{2}{c}{ \textbf{Adversarial}}&\multicolumn{2}{|c}{ \textbf{Average}}
    
\\
\cline{4-15}
    &&&\textbf{F1} &\textbf{BLEU-1} &\textbf{F1} &\textbf{BLEU-1}&\textbf{F1} &\textbf{BLEU-1}&\textbf{F1} &\textbf{BLEU-1}&\textbf{F1} &\textbf{BLEU-1}&\textbf{F1} &\textbf{BLEU-1} \\
    
       \toprule
  \multirow{12}*{\rotatebox{90}{\textbf{Qwen2.5}}} &\multirow{6}*{\rotatebox{90}{\textbf{1.5b}}}  &LCM. &9.05 &6.55& 4.25& 4.04 &9.91 &8.50 &11.15& 8.67 &40.38 &40.23 &14.95 &13.60\\
  &&RA. &6.61 &4.93 &2.55 &2.51 &5.31 &12.24 &10.13 &7.54 &5.42 &27.32 & 6.00&10.91\\
  &&MB. &11.14 &8.25 &4.46 &2.87 &8.05 &6.21 &13.42 &11.01 &36.76 &34.00 &14.77 &12.47\\
  &&MG. &10.44 &7.61 &4.21 &3.89 &13.42 &11.64 &9.56 &7.34 &31.51 &28.90 & 13.83&11.88\\&&AM. &18.23 &11.94 &24.32 &19.74 &16.48 &14.31 &23.63 &19.23 &46.00 &43.26 &25.73 &21.70\\
&& \textbf{Ours} &  \textbf{21.44}&  \textbf{14.24}&  \textbf{32.43}& \textbf{29.76}&  \textbf{19.23}& \textbf{15.37}& \textbf{28.47}& \textbf{21.98}&  \textbf{50.27}& \textbf{49.36}&  \textbf{30.37}& \textbf{26.14}\\

  \cline{2-15}
  &\multirow{6}*{\rotatebox{90}{\textbf{3b}}}&LCM.&4.61 &4.29 &3.11 &2.71 &4.55 &5.97 &7.03 &5.69 &16.95 &14.81 &7.25 &6.69\\
  &&RA. &2.47 &1.78 &3.01 &3.01 &5.57 &5.22 &3.25 &2.51 &15.78 &14.01 &6.02 &5.30\\
  &&MB. &3.60 &3.39 &1.72 &1.97 &6.63 &6.58 &4.11 &3.32 &13.07 &10.30 &5.83 &5.11\\
  &&MG. &5.07 &4.31 &2.94 &2.95 &7.04 &7.10 &7.26 &5.52 &14.47 &12.39 &7.36 &6.45\\
  &&AM. &12.57 &9.01 &27.59 &25.07 &5.33 &5.28 &17.23 &13.12 &27.91 &25.15 &18.13 &15.53\\
 && \textbf{Ours} &  \textbf{18.37}&  \textbf{12.23}&  \textbf{31.25}& \textbf{26.36}&  \textbf{16.23}& \textbf{13.27}& \textbf{24.24}& \textbf{19.24}&  \textbf{38.24}& \textbf{37.24}& \textbf{25.67} & \textbf{21.67}\\
\toprule

  \multirow{12}*{\rotatebox{90}{\textbf{Llama3.2}}} &\multirow{6}*{\rotatebox{90}{\textbf{1.5b}}}  &LCM. &11.25 &9.18 &7.38 &6.82 &11.90 &10.38 &12.86 &10.50 &41.89 &37.27 & 17.06&14.83\\
  &&RA. &5.96 &5.12 &1.93 &2.30 &12.46 &11.17& 7.75 &6.03 &44.64 &40.15 & 14.55&12.95\\
  &&MB. &13.18 &10.03 &7.61 &6.27 &15.78 &12.94 &17.30 &14.03 &52.61 &47.53 &21.30 &18.16\\
  &&MG. &9.19 &6.96 &4.02 &4.79 &11.14 &8.24 &10.16 &7.68 &49.75 &45.11 & 16.85&14.56\\&&AM. &19.06 &11.71 &17.80 &10.28 &17.55 &14.67 &28.51 &24.13 &58.81 &54.28 &28.35 &23.01\\
&& \textbf{Ours} &  \textbf{21.24}&  \textbf{12.34}&  \textbf{25.23}& \textbf{16.23}&  \textbf{18.23}& \textbf{15.23}& \textbf{29.37}& \textbf{27.35}&  \textbf{60.27}& \textbf{57.23}& \textbf{30.87} & \textbf{25.68}\\

  \cline{2-15}
  &\multirow{6}*{\rotatebox{90}{\textbf{3b}}}&LCM.&6.88 &5.77 &4.37 &4.40 &10.65 &9.29 &8.37 &6.93 &30.25 &28.46 &12.10 &10.97\\
  &&RA. &2.47 &1.78 &3.01 &3.01& 5.57 &5.22 &3.25 &2.51 &15.78 &14.01 &6.02 &5.31\\
  &&MB. &6.19 &4.47 &3.49 &3.13 &4.07 &4.57 &7.61 &6.03 &18.65 &17.05 &7.80 &7.05\\
  &&MG. &5.32 &3.99 &2.68 &2.72 &5.64 &5.54 &4.32 &3.51 &21.45 &19.37 &7.88 &7.03\\
  &&AM. &17.44 &11.74 &22.38 &14.24 &12.53 &11.83 &28.14 &23.87 &42.04 &40.60 &24.51 &20.46\\
  && \textbf{Ours} &  \textbf{20.23}&  \textbf{13.24}&  \textbf{24.34}& \textbf{18.24}&  \textbf{17.23}& \textbf{13.27}& \textbf{29.37}& \textbf{24.34}&  \textbf{52.34}& \textbf{47.34}&  \textbf{28.70}& \textbf{23.29}\\
  \toprule

  \multirow{12}*{\rotatebox{90}{\textbf{DeepSeek-R1}}} &\multirow{6}*{\rotatebox{90}{\textbf{1.5b}}}  &LCM. & 21.43&16.81 &17.78 & 14.77&11.98&10.00 & 31.22&27.74 & 41.34& 35.23 &24.75 &20.91\\
  &&RA. & 7.11&5.67 &14.90 &9.23 & 4.37&4.25 &8.98 & 6.78&8.23& 7.11&8.72 &6.61\\
  &&MB. &5.42 &5.11 &9.77 &8.24 &5.11 &4.12 &7.18 &7.10 &7.76& 6.00&7.05 &6.11\\
  &&MG. & 24.67&21.12 &24.23 &18.76 &8.24 &7.23 &40.42 &37.76& 42.3&41.44  & 27.97&25.26\\&&AM. &18.88 &13.47 &39.24 &35.23 &7.23 &7.10 &31.12 &26.34 &30.21 &29.34 & 25.33&22.30\\
&& \textbf{Ours} &  \textbf{26.37}&  \textbf{25.67}&  \textbf{39.45}& \textbf{38.57}&  \textbf{20.78}& \textbf{15.23}& \textbf{43.98}& \textbf{38.29}&  \textbf{63.30}& \textbf{59.32}& \textbf{38.78} & \textbf{35.42}\\

  \cline{2-15}
  &\multirow{6}*{\rotatebox{90}{\textbf{7b}}}&LCM.&16.13 & 17.24&10.2 &6.23 &15.27 &15.3 &47.24 & 46.1& 40.00& 35.23&25.77 &24.02\\
  &&RA. & 13.31& 9.97&4.47 & 3.01& 7.78& 5.23& 11.23& 9.78& 9.81& 8.03&9.32 &7.20\\
  &&MB. &5.12 &5.34 &3.57 &2.47 &7.34 &8.27 & 6.54& 8.11&6.23 & 5.24& 5.76&5.89\\
  &&MG. &26.78 &21.67 &18.23 & 12.35&12.27 &11.87 & \textbf{53.24}&\textbf{49.34} &33.23 & 31.25&28.75&25.30\\
  &&AM. & 31.24& 21.34&34.27 & 29.34&16.87 &15.89 &45.24 &41.23 &30.24 &30.11 & 31.57&27.58\\
  && \textbf{Ours} &  \textbf{34.23}&  \textbf{24.34}&  \textbf{38.67}& \textbf{37.36}&  \textbf{21.55}& \textbf{17.21}& 42.34& 39.47&  \textbf{60.34}& \textbf{55.34}& \textbf{39.43} & \textbf{34.74}\\

    \toprule
\end{tabular}

 \caption{\textbf{Experimental results on the LoCoMo dataset are reported across five QA task categories.} We evaluate multiple methods using F1 and BLEU-1 scores (in \%). The best performance in each category is highlighted in bold, while our proposed method, H-MEM (shaded in gray in the table), consistently demonstrates competitive or superior performance across six foundation language models.}
\label{tab:table1}
\end{table*}

\subsection{Results and Analysis}

\noindent \textbf{Comparison to Baselines. }As shown in the Table \ref{tab:table1}, we conducted systematic comparative experiments on five categories of tasks from the LoCoMo dataset, evaluating five mainstream baseline methods across multiple large language models (LLMs) of varying scales. This comprehensive evaluation aims to assess the effectiveness and stability of H-MEM in long-term dialogue tasks. Overall, our method consistently achieves the highest average F1 and BLEU-1 scores across all model and task configurations, with improvements of 14.98 and 12.77 points over the baselines, respectively, demonstrating its generalizability and significant advantages. In relatively basic tasks such as Single-Hop, Temporal, and Open-Domain dialogue, H-MEM consistently outperforms all baselines, providing preliminary evidence of its effectiveness in long-term dialogue modeling. More notably, in more challenging tasks such as Multi-Hop and Adversarial dialogues—which demand stronger long-range dependency modeling and complex reasoning—our method exhibits outstanding performance. Specifically, in the Multi-Hop task, H-MEM outperforms baselines by an average of 21.25 and 17.65 points in F1 and BLEU-1 scores, respectively; in the Adversarial task, it achieves gains of 16.71 and 12.03 points, respectively. These substantial improvements further validate H-MEM's capability in long-term memory storage and efficient retrieval. Additionally, our method consistently maintains leading performance across models of different scales (including 1.5B, 3B, and 7B), indicating strong model-agnosticism and cross-architecture generalizability. Notably, H-MEM achieves significant performance gains even in smaller models (e.g., 1.5B), suggesting its practicality and applicability in resource-constrained scenarios.\\
\\

\begin{figure}[t]
    \centering
    \includegraphics[width=1\linewidth]{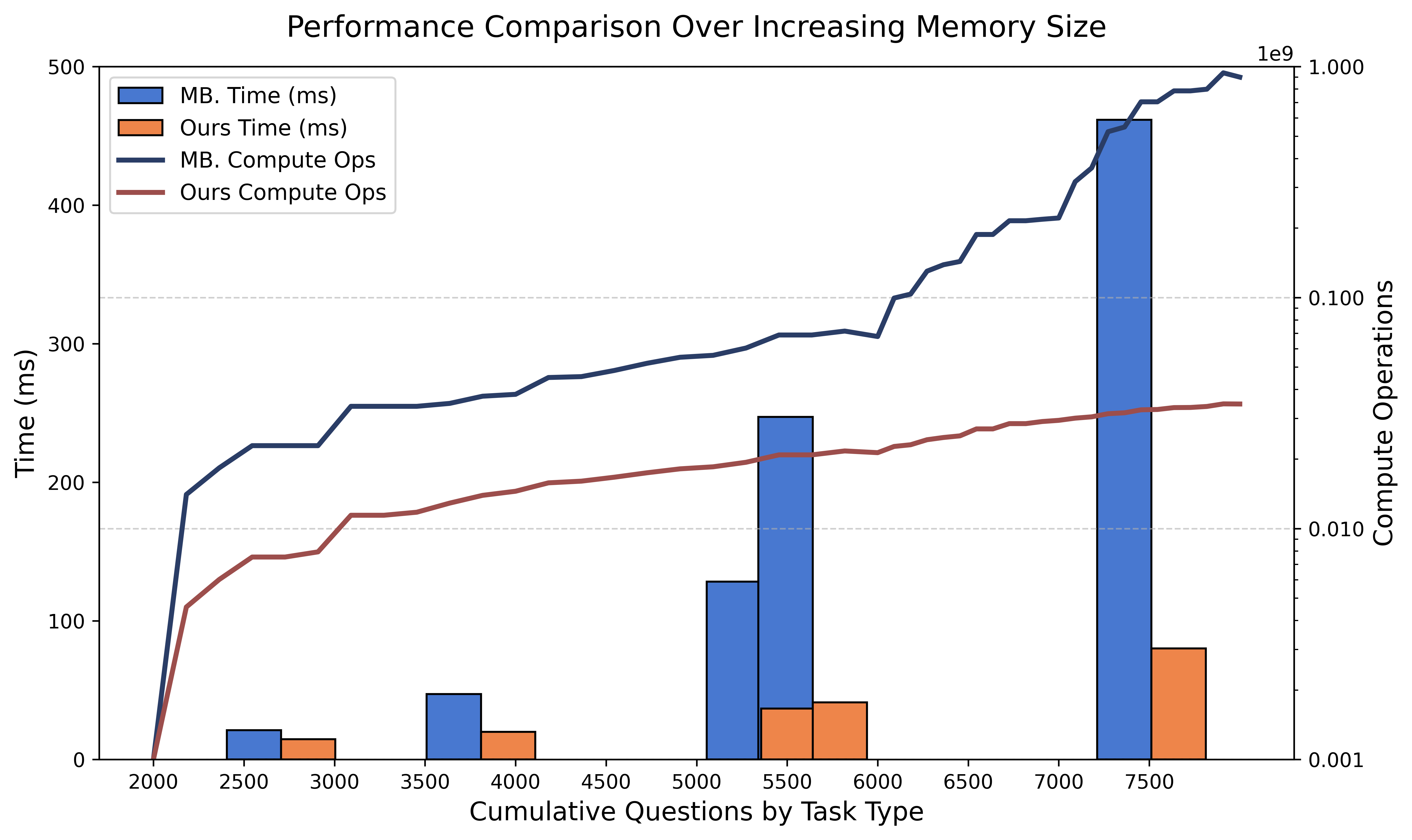}
    \caption{\textbf{Comparative analysis of computational efficiency. }We compare the calculation amount and time of H-MEM and baseline (MemoryBank) when using Qwen-1.5b to perform five types of QA tasks to verify the efficiency of H-MEM's memory retrieval. We select the end of each task type as a checkpoint to calculate the calculation time of completing a task type, and calculate the calculation amount every ten tasks.}
    \label{fig:calculate}
\end{figure}

\noindent \textbf{Computational Efficiency Analysis. }To evaluate the efficiency of the hierarchical memory storage and retrieval mechanism of H-MEM, we compare its performance with the baseline under conditions of large-scale memory and substantial irrelevant memory interference, as shown in Table \ref{tab:efficiency}. Specifically, we assess both the computational cost and latency during memory retrieval, as well as the quality and accuracy of the generated answers. In terms of latency, the H-MEM inference time remains below 100 ms, even at maximum memory load, while the baseline exceeds 400ms, making it 5 times slower than H-MEM. Despite this substantial reduction in computation and latency, H-MEM consistently outperforms the baseline in answer quality and accuracy, particularly in long-context dialogue settings. As shown in the Figure \ref{fig:calculate}, the calculation amount of baseline shows an almost exponential growth trend with the accumulation of memory, while H-MEM still shows a slow increase and gradually stabilizes trend with the increasing memory amount. These results demonstrate that the efficiency advantages of H-MEM become increasingly pronounced as the memory size grows.\\
\\

\begin{table}[t]
    \centering
 \scriptsize
  \begin{tabular}{c|c|cccc}
    \toprule  
   \multirow{1}*{\textbf{Task}} & \multirow{1}*{\textbf{Method}} & \textbf{F1}& \textbf{BLEU-1}
    & \textbf{Compute Ops} &\textbf{Time (ms)} 
    
       \\
    
       \toprule
  \multirow{2}*{SH.}&MB.&12.45&8.25&$3.81\times 10^7$&21.21\\

                &Ours&21.44&14.24&$1.45\times 10^7$&14.55\\

    \toprule
     \multirow{2}*{MH.}& MB.&4.45&3.92& $6.78\times 10^7$&47.22\\

                &Ours&31.24&27.34& $2.13\times 10^7$&19.88\\

    \toprule

    \multirow{2}*{T.}& MB.&6.32&6.12&$2.21\times 10^8$&128.34\\

                 &Ours&18.28&14.14&$2.94\times 10^7$&36.74\\
    
    \toprule
    \multirow{2}*{OD.}& MB.&9.36&4.23&$9.00\times 10^8$&247.28\\

                 &Ours&27.28&19.37&$3.46\times 10^7$&41.27\\
    \toprule
    \multirow{2}*{A.}& MB.&20.13&17.36&$7.34\times 10^9$&461.54\\

                 &Ours&43.23&40.12&$4.38\times 10^7$&80.07\\
    \toprule

\end{tabular}
  
 \caption{\textbf{Additional computational efficiency comparison experiment.} In the case of increasing memory storage capacity, the computational requirements for H-MEM and MemoryBank retrieval memory are compared. }
 \label{tab:efficiency}
\end{table}

\begin{figure}[t]
    \centering
    \includegraphics[width=0.9\linewidth]{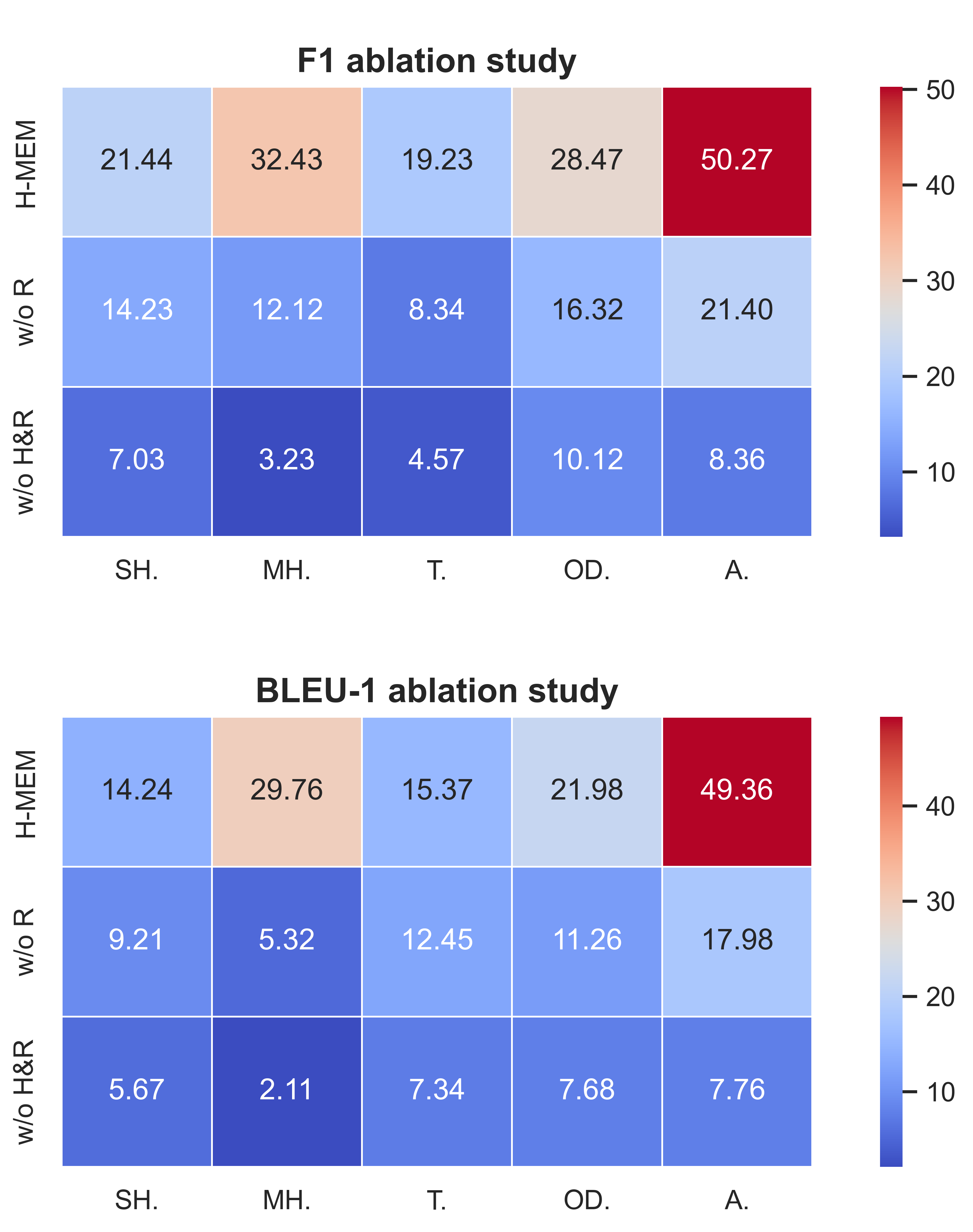}
    \caption{\textbf{Ablation study result. }In this figure, H represents the hierarchical memory storage of H-MEM, and R is the position index retrieval in H-MEM.}
    \label{fig:ablation}
\end{figure}

\noindent \textbf{Ablation Study. }To evaluate the effectiveness of the H-MEM module in memory storage and retrieval, we conducted ablation studies on multiple benchmark tasks using the Qwen-1.5B model. As illustrated in the Figure \ref{fig:ablation}, we compared three configurations: removing the memory retrieval component of H-MEM (w/o R.), removing both the hierarchical memory storage and retrieval mechanisms (w/o H\&R.), and the full H-MEM module. The experimental results demonstrate a clear performance degradation in long-term dialogue tasks as key components of H-MEM are progressively removed. These findings indicate that: (1) the memory retrieval mechanism of H-MEM is ineffective without the support of structured hierarchical memory storage; and (2) the synergy between memory storage and retrieval is essential for enabling LLM agents to perform well in long-term conversational settings.

\section{Conclusion and Future Work}

We propose a Hierarchical Memory (H-MEM) architecture for LLM Agents that organizes and updates memory in a multi-level fashion based on the degree of semantic abstraction. Using a four-layer memory layer that summarizes layer by layer, and adding a position index corresponding to the sub-memory at the end, to achieve structured memory storage and efficient memory retrieval. To evaluate the effectiveness of the proposed H-MEM system, we conducted comparative experiments against five mainstream baseline methods across five types of long-term dialogue question-answering tasks. Experimental results demonstrate that H-MEM exhibits significant advantages in preserving and accessing long-term conversational memory. Furthermore, computational efficiency experiments confirm that H-MEM achieves high retrieval efficiency even under complex memory conditions. Future work will focus on developing more efficient memory mechanisms and extending H-MEM to support multimodal memory representations.

\section{Limitations}

\noindent \textbf{Insufficient Support for Multi-Modal Memory. }The current H-MEM architecture primarily focuses on text-based memory storage and retrieval, with limited support for multimodal memory. In practical applications, interactions between users and LLM agents may involve multiple modalities, such as images, audio, and video. However, the memory storage and retrieval mechanisms of H-MEM have not yet fully considered the integration of these multimodal information sources. For example, in conversations that include images or videos, H-MEM is unable to directly process non-textual information, thereby limiting its applicability in multimodal dialogue scenarios.\\
\\

\noindent \textbf{Memory Capacity Limitations. }Despite the significant improvement in memory retrieval efficiency achieved through its hierarchical structure, the memory capacity of H-MEM is still limited. As the dialogue continues and the volume of memory content increases, the storage space of H-MEM may gradually become exhausted. Although external storage devices can be used to expand memory capacity, this may introduce additional latency and management overhead. Moreover, with the increase in memory capacity, effectively managing the memory lifecycle (such as memory expiration and deletion) becomes an issue that needs to be addressed.\\
\\

\noindent \textbf{User Privacy and Security Concerns. }In long-term dialogues, H-MEM stores a large amount of user interaction information, which may involve users' privacy and sensitive data. Ensuring the secure storage and access of this memory content is an important issue. For example, users may not want certain personal information to be stored long-term or used for other purposes. Therefore, it is necessary to design effective privacy protection mechanisms to restrict access to and use of memory. Additionally, as the volume of memory content increases, preventing malicious attackers from tampering with or stealing memory data is also a security issue that needs to be considered.

\bibliography{main}

\begin{thebibliography}{23}
\providecommand{\natexlab}[1]{#1}

\bibitem[{DeepSeek-AI(2025)}]{deepseek-r1}
DeepSeek-AI. 2025.
\newblock \href {https://arxiv.org/abs/2501.12948} {Deepseek-r1: Incentivizing reasoning capability in llms via reinforcement learning}.
\newblock \emph{Preprint}, arXiv:2501.12948.

\bibitem[{Douze et~al.(2024)Douze, Guzhva, Deng, Johnson, Szilvasy, Mazaré, Lomeli, Hosseini, and Jégou}]{faiss}
Matthijs Douze, Alexandr Guzhva, Chengqi Deng, Jeff Johnson, Gergely Szilvasy, Pierre-Emmanuel Mazaré, Maria Lomeli, Lucas Hosseini, and Hervé Jégou. 2024.
\newblock \href {https://arxiv.org/abs/2401.08281} {The faiss library}.

\bibitem[{Gu et~al.(2024)Gu, Jiang, Shi, Tan, Zhai, Xu, Li, Shen, Ma, Liu et~al.}]{LLMsurvey2}
Jiawei Gu, Xuhui Jiang, Zhichao Shi, Hexiang Tan, Xuehao Zhai, Chengjin Xu, Wei Li, Yinghan Shen, Shengjie Ma, Honghao Liu, and 1 others. 2024.
\newblock A survey on llm-as-a-judge.
\newblock \emph{arXiv preprint arXiv:2411.15594}.

\bibitem[{Hatalis et~al.(2023)Hatalis, Christou, Myers, Jones, Lambert, Amos-Binks, Dannenhauer, and Dannenhauer}]{needmemory}
Kostas Hatalis, Despina Christou, Joshua Myers, Steven Jones, Keith Lambert, Adam Amos-Binks, Zohreh Dannenhauer, and Dustin Dannenhauer. 2023.
\newblock Memory matters: The need to improve long-term memory in llm-agents.
\newblock In \emph{Proceedings of the AAAI Symposium Series}, volume~2, pages 277--280.

\bibitem[{Huang et~al.(2024{\natexlab{a}})Huang, Liu, Chen, Wang, Wang, Lian, Wang, Tang, and Chen}]{agentsurvey}
Xu~Huang, Weiwen Liu, Xiaolong Chen, Xingmei Wang, Hao Wang, Defu Lian, Yasheng Wang, Ruiming Tang, and Enhong Chen. 2024{\natexlab{a}}.
\newblock Understanding the planning of llm agents: A survey.
\newblock \emph{arXiv preprint arXiv:2402.02716}.

\bibitem[{Huang et~al.(2024{\natexlab{b}})Huang, Liu, Chen, Wang, Wang, Lian, Wang, Tang, and Chen}]{LLMsurvey6}
Xu~Huang, Weiwen Liu, Xiaolong Chen, Xingmei Wang, Hao Wang, Defu Lian, Yasheng Wang, Ruiming Tang, and Enhong Chen. 2024{\natexlab{b}}.
\newblock Understanding the planning of llm agents: A survey.
\newblock \emph{arXiv preprint arXiv:2402.02716}.

\bibitem[{Lee et~al.(2024)Lee, Chen, Furuta, Canny, and Fischer}]{readingagent}
Kuang-Huei Lee, Xinyun Chen, Hiroki Furuta, John Canny, and Ian Fischer. 2024.
\newblock \href {https://arxiv.org/abs/2402.09727} {A human-inspired reading agent with gist memory of very long contexts}.
\newblock \emph{Preprint}, arXiv:2402.09727.

\bibitem[{Li et~al.(2024)Li, Wen, Wang, Li, Yuan, Liu, Liu, Xu, Wang, Sun et~al.}]{agentsurvey2}
Yuanchun Li, Hao Wen, Weijun Wang, Xiangyu Li, Yizhen Yuan, Guohong Liu, Jiacheng Liu, Wenxing Xu, Xiang Wang, Yi~Sun, and 1 others. 2024.
\newblock Personal llm agents: Insights and survey about the capability, efficiency and security.
\newblock \emph{arXiv preprint arXiv:2401.05459}.

\bibitem[{Maharana et~al.(2024)Maharana, Lee, Tulyakov, Bansal, Barbieri, and Fang}]{LoCoMoDataset}
Adyasha Maharana, Dong-Ho Lee, Sergey Tulyakov, Mohit Bansal, Francesco Barbieri, and Yuwei Fang. 2024.
\newblock Evaluating very long-term conversational memory of llm agents.
\newblock \emph{arXiv preprint arXiv:2402.17753}.

\bibitem[{Packer et~al.(2023)Packer, Fang, Patil, Lin, Wooders, and Gonzalez}]{memgpt}
Charles Packer, Vivian Fang, Shishir\_G Patil, Kevin Lin, Sarah Wooders, and Joseph\_E Gonzalez. 2023.
\newblock Memgpt: Towards llms as operating systems.

\bibitem[{Salama et~al.(2025)Salama, Cai, Yuan, Currey, Sunkara, Zhang, and Benajiba}]{meminsight}
Rana Salama, Jason Cai, Michelle Yuan, Anna Currey, Monica Sunkara, Yi~Zhang, and Yassine Benajiba. 2025.
\newblock Meminsight: Autonomous memory augmentation for llm agents.
\newblock \emph{arXiv preprint arXiv:2503.21760}.

\bibitem[{Shen(2024)}]{LLMsurvey3}
Zhuocheng Shen. 2024.
\newblock Llm with tools: A survey.
\newblock \emph{arXiv preprint arXiv:2409.18807}.

\bibitem[{Touvron et~al.(2023)Touvron, Lavril, Izacard, Martinet, Lachaux, Lacroix, Rozière, Goyal, Hambro, Azhar, Rodriguez, Joulin, Grave, and Lample}]{llama}
Hugo Touvron, Thibaut Lavril, Gautier Izacard, Xavier Martinet, Marie-Anne Lachaux, Timothée Lacroix, Baptiste Rozière, Naman Goyal, Eric Hambro, Faisal Azhar, Aurelien Rodriguez, Armand Joulin, Edouard Grave, and Guillaume Lample. 2023.
\newblock \href {https://arxiv.org/abs/2302.13971} {Llama: Open and efficient foundation language models}.
\newblock \emph{Preprint}, arXiv:2302.13971.

\bibitem[{Wang et~al.(2023)Wang, Liang, Yang, Huang, Wu, Wu, Lu, Ma, and Li}]{SCM}
Bing Wang, Xinnian Liang, Jian Yang, Hui Huang, Shuangzhi Wu, Peihao Wu, Lu~Lu, Zejun Ma, and Zhoujun Li. 2023.
\newblock Enhancing large language model with self-controlled memory framework.
\newblock \emph{arXiv preprint arXiv:2304.13343}.

\bibitem[{Wu et~al.(2025)Wu, Yang, Zhan, Yuan, Chao, and Wong}]{LLMsurvey5}
Junchao Wu, Shu Yang, Runzhe Zhan, Yulin Yuan, Lidia~Sam Chao, and Derek~Fai Wong. 2025.
\newblock A survey on llm-generated text detection: Necessity, methods, and future directions.
\newblock \emph{Computational Linguistics}, pages 1--66.

\bibitem[{Xu et~al.(2025)Xu, Liang, Mei, Gao, Tan, and Zhang}]{Amem}
Wujiang Xu, Zujie Liang, Kai Mei, Hang Gao, Juntao Tan, and Yongfeng Zhang. 2025.
\newblock A-mem: Agentic memory for llm agents.
\newblock \emph{arXiv preprint arXiv:2502.12110}.

\bibitem[{Yang et~al.(2024{\natexlab{a}})Yang, Yang, Hui, Zheng, Yu, Zhou, Li, Li, Liu, Huang, Dong, Wei, Lin, Tang, Wang, Yang, Tu, Zhang, Ma, Xu, Zhou, Bai, He, Lin, Dang, Lu, Chen, Yang, Li, Xue, Ni, Zhang, Wang, Peng, Men, Gao, Lin, Wang, Bai, Tan, Zhu, Li, Liu, Ge, Deng, Zhou, Ren, Zhang, Wei, Ren, Fan, Yao, Zhang, Wan, Chu, Liu, Cui, Zhang, and Fan}]{qwen2}
An~Yang, Baosong Yang, Binyuan Hui, Bo~Zheng, Bowen Yu, Chang Zhou, Chengpeng Li, Chengyuan Li, Dayiheng Liu, Fei Huang, Guanting Dong, Haoran Wei, Huan Lin, Jialong Tang, Jialin Wang, Jian Yang, Jianhong Tu, Jianwei Zhang, Jianxin Ma, and 40 others. 2024{\natexlab{a}}.
\newblock Qwen2 technical report.
\newblock \emph{arXiv preprint arXiv:2407.10671}.

\bibitem[{Yang et~al.(2024{\natexlab{b}})Yang, Yang, Zhang, Hui, Zheng, Yu, Li, Liu, Huang, Wei, Lin, Yang, Tu, Zhang, Yang, Yang, Zhou, Lin, Dang, Lu, Bao, Yang, Yu, Li, Xue, Zhang, Zhu, Men, Lin, Li, Xia, Ren, Ren, Fan, Su, Zhang, Wan, Liu, Cui, Zhang, and Qiu}]{qwen2.5}
An~Yang, Baosong Yang, Beichen Zhang, Binyuan Hui, Bo~Zheng, Bowen Yu, Chengyuan Li, Dayiheng Liu, Fei Huang, Haoran Wei, Huan Lin, Jian Yang, Jianhong Tu, Jianwei Zhang, Jianxin Yang, Jiaxi Yang, Jingren Zhou, Junyang Lin, Kai Dang, and 22 others. 2024{\natexlab{b}}.
\newblock Qwen2.5 technical report.
\newblock \emph{arXiv preprint arXiv:2412.15115}.

\bibitem[{Yao et~al.(2024)Yao, Duan, Xu, Cai, Sun, and Zhang}]{LLMsurvey1}
Yifan Yao, Jinhao Duan, Kaidi Xu, Yuanfang Cai, Zhibo Sun, and Yue Zhang. 2024.
\newblock A survey on large language model (llm) security and privacy: The good, the bad, and the ugly.
\newblock \emph{High-Confidence Computing}, page 100211.

\bibitem[{Yi et~al.(2024)Yi, Ouyang, Liu, Liao, Xu, and Shen}]{multidialogue}
Zihao Yi, Jiarui Ouyang, Yuwen Liu, Tianhao Liao, Zhe Xu, and Ying Shen. 2024.
\newblock A survey on recent advances in llm-based multi-turn dialogue systems.
\newblock \emph{arXiv preprint arXiv:2402.18013}.

\bibitem[{Zhang et~al.(2024{\natexlab{a}})Zhang, Bo, Ma, Li, Chen, Dai, Zhu, Dong, and Wen}]{memorysurvey}
Zeyu Zhang, Xiaohe Bo, Chen Ma, Rui Li, Xu~Chen, Quanyu Dai, Jieming Zhu, Zhenhua Dong, and Ji-Rong Wen. 2024{\natexlab{a}}.
\newblock A survey on the memory mechanism of large language model based agents.
\newblock \emph{arXiv preprint arXiv:2404.13501}.

\bibitem[{Zhang et~al.(2024{\natexlab{b}})Zhang, Bo, Ma, Li, Chen, Dai, Zhu, Dong, and Wen}]{memorysurvey1}
Zeyu Zhang, Xiaohe Bo, Chen Ma, Rui Li, Xu~Chen, Quanyu Dai, Jieming Zhu, Zhenhua Dong, and Ji-Rong Wen. 2024{\natexlab{b}}.
\newblock A survey on the memory mechanism of large language model based agents.
\newblock \emph{arXiv preprint arXiv:2404.13501}.

\bibitem[{Zhong et~al.(2024)Zhong, Guo, Gao, Ye, and Wang}]{memorybank}
Wanjun Zhong, Lianghong Guo, Qiqi Gao, He~Ye, and Yanlin Wang. 2024.
\newblock Memorybank: Enhancing large language models with long-term memory.
\newblock In \emph{Proceedings of the AAAI Conference on Artificial Intelligence}, volume~38, pages 19724--19731.

\end{thebibliography}



\end{document}